\documentclass[journal,12pt,draftclsnofoot,onecolumn,letterpaper,doublespace]{IEEEtran}

\usepackage{cite,graphicx,subfig,multirow,array}

\usepackage{amsfonts,amssymb}
\usepackage[cmex10]{amsmath}
\usepackage{algorithmic}
\usepackage[ruled]{algorithm2e}	
\usepackage{epstopdf}
\usepackage{hyperref}

\usepackage[usenames,dvipsnames]{color}
\usepackage{soul}

\usepackage{colortbl}
\definecolor{mygray}{gray}{.9}

\begin{document} 

\pagenumbering{gobble}

\title{Light Field Compression \\with Disparity Guided Sparse Coding \\based on Structural Key Views}

\author{\IEEEauthorblockN{Jie~Chen, Junhui~Hou, Lap-Pui~Chau}

\thanks{J. Chen and L.-P. Chau are with the School of Electrical \& Electronic Engineering, Nanyang Technological University, Singapore (e-mail: \{Chen.Jie, ELPChau\}@ntu.edu.sg), J. Hou is with the Department of Computer Science, City University of Hong Kong (e-mail: jh.hou@cityu.edu.hk).}
}


\maketitle

\begin{abstract}
	
Recent imaging technologies are rapidly evolving for sampling richer and more immersive representations of the 3D world. And one of the emerging technologies are light field (LF) cameras based on micro-lens arrays. To record the directional information of the light rays, a much larger storage space and transmission bandwidth are required by a LF image as compared with a conventional 2D image of similar spatial dimension, and the compression of LF data becomes a vital part of its application. 

In this paper, we propose a LF codec that fully exploits the intrinsic geometry between the LF sub-views by first approximating the LF with disparity guided sparse coding over a perspective shifted light field dictionary. The sparse coding is only based on several optimized Structural Key Views (SKV); however the entire LF can be recovered from the coding coefficients. By keeping the approximation identical between encoder and decoder, only the residuals of the non-key views, disparity map and the SKVs need to be compressed into the bit stream. An optimized SKV selection method is proposed such that most LF spatial information could be preserved. And to achieve optimum dictionary efficiency, the LF is divided into several Coding Regions (CR), over which the reconstruction works individually. Experiments and comparisons have been carried out over benchmark LF dataset, which show that the proposed \textit{SC-SKV} codec produces convincing compression results in terms of both rate-distortion performance and visual quality compared with \textit{High Efficiency Video Coding} (\textit{HEVC}): with 47.87\% BD-rate reduction and 1.59 dB BD-PSNR improvement achieved on average, especially with up to 4 dB improvement for low bit rate scenarios.

\end{abstract}
\IEEEpeerreviewmaketitle

\section{Introduction} \label{sec:intro}

Recent imaging technologies are rapidly evolving for sampling richer and more immersive representations of the 3D world. And one of the emerging technologies are light field (LF) cameras based on micro-lens arrays \cite{lippmann1908la} \cite{Ng2005}. Light field imaging revolutionized traditional photography by providing many new exciting functionalities such as after-shot image manipulations, i.e., refocusing and arbitrary view synthesis. The Depth and geometric information that can be derived from a LF image, can be extremely beneficial in image processing applications such as segmentation, salient object detection \cite{li2014saliency}, and action recognition.

\begin{figure}[!t]
	\centering
	\includegraphics[width=3.5in]{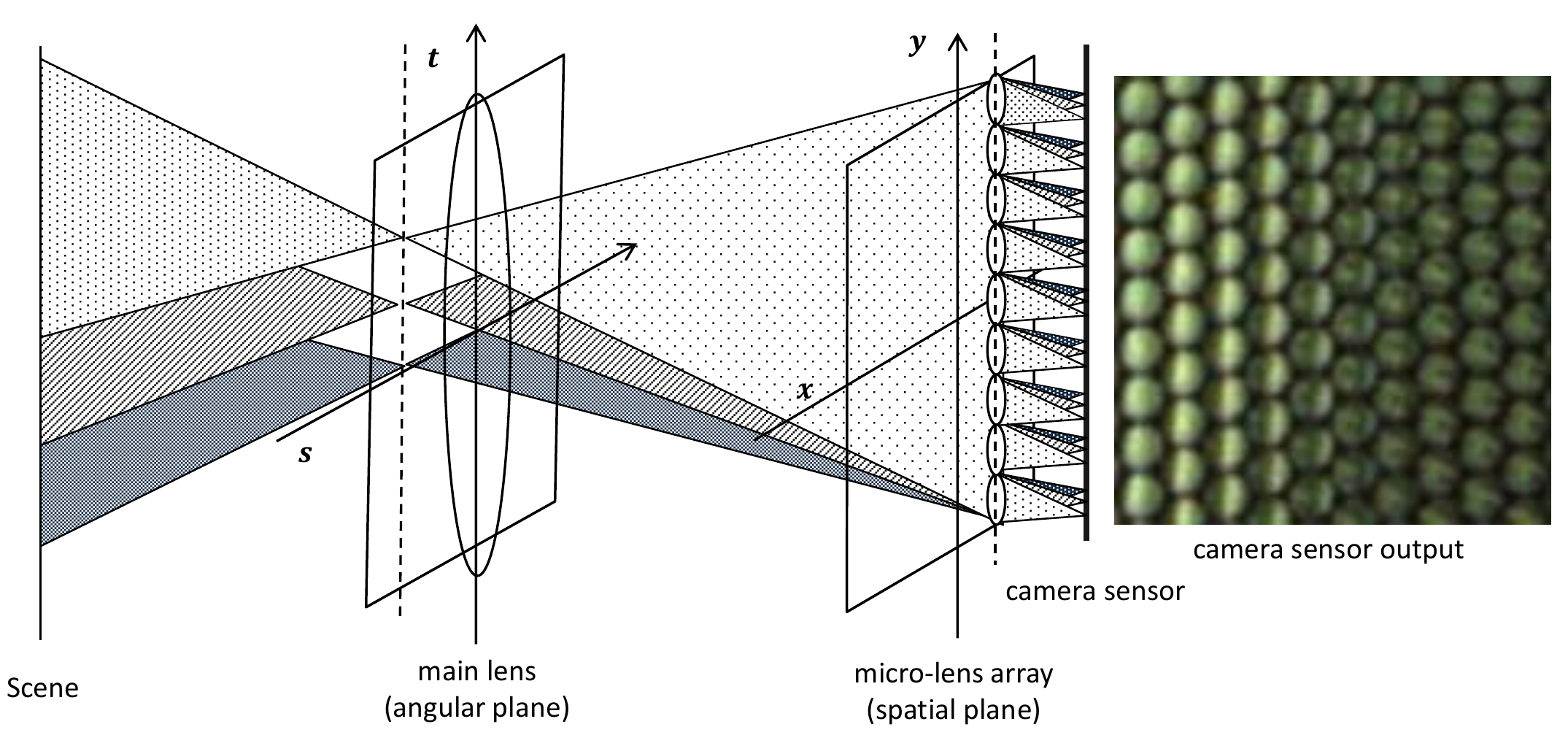}
	\caption{Structure of a micro-lens based LF camera, and sample of the camera sensor data.}
	\label{fig_plenopticCamera}
\end{figure}

The light field is a vector function that describes the amount of light propagating in every direction through every point in space \cite{lippmann1908la}. As illustrated in Fig. \ref{fig_plenopticCamera}, light field is usually represented as four-dimensional data: $L(x,y,s,t)$, in which $(x,y)$ is at the micro-lens array plane (which is near the focal plane of the main lens); and $(s,t)$ is at the camera main lens plane. The four parameters at two planes can sufficiently describe the propagation of all light rays in the imaging system. 

Numerous brands and prototypes of light field (LF) cameras have been designed and developed over the years, its core technology can be divided into three categories: multi-camera array based \cite{wilburn2004high,taguchi2010axial}, spatial modulation based \cite {Nagahara2010, ashok2010, babacan2012, marwah2013compressive, chen2015light}, and micro-lens array based cameras \cite{Ng2005}. Different cameras have their unique structural design and respective advantages, either in spatial-angular resolution, angular baseline, or equipment cost. In recent years, plenoptic cameras have gradually dominated the consumer LF acquisition market because of its portability and low-cost. Fig. \ref{fig_plenopticCamera} shows the typical structure of a plenoptic camera. With the inserted micro-lens array, different light rays that pass through a certain point on the main lens are redirected to similar directions by the micro-lenses; And vice versa, different light ray directions behind the micro-lenses correspond to different viewing angles from the main lens plane. The final output lenselet image is shown at the right of Fig. \ref{fig_plenopticCamera}. It requires further decoding before it could be used for various applications. The decoding is directly related to camera calibration parameters such as micro-lens center spacing, and lens line rotation angle.

The high dimensionality of the LF data makes each light field capture extremely large in size. For a commercial LF camera with a sensor of 5000x7000 (35 million) pixels, each acquisition can only produce spatial views with VGA resolution if without super-resolution. However the file size will be mroe than 80 MBytes. Therefore with higher market demand for high-resolution LF images, and hopefully even for LF videos, LF data compression becomes a vital issue.

\subsection{Motivation}

The decoded 4D light field $L(x,y,s,t)$ can be considered as a 2D sequence of sub-view images (SVI). Each sub-view index pair $(s,t)$ corresponds to a unique viewing angle. Strong correlation exists in $L$ not just in the 2D spatial domain ($x,y$), but also in the angular domain $(s,t)$. To efficiently remove the correlation in all four dimensions, the particular geometry and structure of the LF data has to be studied carefully.

Based on the format of LF data, current LF compression methods can be roughly divided into two categories: micro-lens array image (MLI) based methods and pseudo sequence based methods. For the first category, MLIs (as shown on the right of Fig. \ref{fig_plenopticCamera}) are directly used for compression. The lenselet image itself is a regular grid of macro-pixel images (MAI). A MAI is formed by the group of pixels under a single micro-lens. Since each pixel in a MAI represents a separate viewing angle, therefore the correlation within each MAI actually represents angular correlation. Recent LF compression methods proposed by Li et al. \cite{li2016scalable} and Conti et al. \cite{conti2016comm,conti2016hevc} both work on MLIs. Li et al. \cite{li2016scalable} attempted to remove the angular correlation via disparity guided view interpolation based on a sparse set of sub-sampled angular examples. Conti et al. \cite{conti2016comm,conti2016hevc} proposed the self-similarity concept to compensate repeated MAI patterns. Both methods can efficiently remove the angular correlation in the LF. However based on the MLI structure of unfocused LF cameras (such as Lytro), the spatial pixel neighbors from a certain SVI are interleaved across different MAIs: i.e., each pixel under the same micro-lens contributes to different SVIs. Such structure makes the spatial correlation more difficult to be removed than angular correlation. Compression methods based on the MLI data format generally work better for MLIs from focused light field cameras \cite{georgiev2010focused}. Unlike unfocused LF cameras, focused LF cameras render a group of pixels under each MAI to one specific SVI. Consequently the spatial pixel neighbors are positioned closer and more visually coherent, which makes the spatial correlation easier to be removed as compared to MLIs from focused LF cameras.
	
Another limitation for MLI based compression methods is that camera specific parameters need to be transmitted to users for each LF image before they can be decoded and used for subsequent applications. This will add extra load into the compressed data stream.

The second category of LF compression methods work on decoded light field data. Angular dimension $(s,t)$ in $L$ will be considered as the time axis in a video sequence, and hence the name ``pseudo sequence''. In this data format, both spatial and angular dimensions are directly available for manipulation. Recently several works have been published \cite{liu2016pseudo, dai2015lenselet, zhao2016light} that investigate how to rearrange the orders of the SVIs in the pseudo sequence to produce better compression results. 

For most recently proposed state-of-the-art LF compression methods, video codecs have been tailored/incorporated into their framework. Video codecs such as the \textit{High Efficiency Video Coding} (\textit{HEVC}) has very flexible prediction modes and partition patterns to host new coding tools, and the \textit{HEVC} test module \cite{hevc2017test} can provide a switch between different modes that achieves the best compression in the sense of rate-distortion optimization (RDO). For example, \textit{HEVC Intra} is used as a competing mode in both \cite{li2016scalable} and \cite{conti2016hevc} to guarantee the performance of their proposed codecs. In \cite{liu2016pseudo}, the JEM software \cite{jem2017jem}, which represents the ongoing effort to further improve compression efficiency over \textit{HEVC}, is used for the compression of LF pseudo sequences after SVI reordering and prediction hierarchy assignment.

We argue that direct use of a video codec for the compression of LF SVI pseudo sequence data is sub-optimal. Compared with normal video sequences, the LF SVI sequences have the following unique characteristics:
\begin{enumerate}
\item Unlike video sequences, the LF SVI sequences are captured at the same time instance. This trivializes the scene temporal dynamics problems (such as camera and object motion, illumination change etc) for motion estimation among the frames.
\item Unlike video sequences where the camera position could be arbitrarily changed along the time axis, the camera viewing positions for the SVIs are fixed with strict linear configurations. This makes the disparity prediction for each single pixel on the SVIs much easier and more precise, compared with the estimation of motion vectors between video frames.
\item Unlike frames in a video sequence where a set of motion vectors need to be calculated and saved for each frame, only one set of ``motion vector'' is enough (which is the disparity map) for the description of pixel displacement in all the SVIs (depending on each SVI's relative location with respect to the center view, refer to Sec. \ref{sec_structure} for detail). 
\end{enumerate}

Based on the above characteristics, a pixel-wise disparity map (motion vectors) can be efficiently calculated for the LF, and just one disparity map will be able to describe the parallax of all pixels across all SVIs. This implies that higher compression performance could be achieved if the disparity map can be used as global guide among all SVIs for the exploitation of angular correlation.

For video codecs such as \textit{HEVC}, the basic processing unit is a Coding Block (CB). It searches spatially and angularly for similar references and removes the correlation on a block basis. This mechanism is very efficient for the compression of normal video frames: since the scene structure and the temporal dynamics among the frames are unknown, pixels in the CB could provide contextual clues for motion estimation, and at the same time it saves storage spaces for motion vectors and reduces computation complexity. For a LF SVI sequence, since the disparity map for all pixels across the sequence can be estimated easily, and given the fact that each pixel within a CB could have different disparities, we believe a LF compression framework will benefit from a pixel based processing structure, with estimated scene disparity between the SVIs as guidance. The work by Li et al. \cite{li2016scalable} is a pixel based method, where the full MLI is predicted on a pixel basis from a sparse set of sub-sampled angular views via disparity guided shifting. However we believe such simple interpolation and reconstruction techniques they used greatly limited the sub-sampling rate (1:3), as well as the final reconstruction quality.
	


\subsection{Our Contribution}

In this paper, we propose a LF codec with disparity guided Sparse Coding over a learned perspective-shifted LF dictionary based on selected Structural Key Views (\textit{SC-SKV}). The contributions of this work can be generalized in the following aspects:

\begin{enumerate}
\item A Coding Region (CR) segmentation method is proposed which divides the entire LF into several separate processing units to achieve optimum dictionary learning and sparse coding efficiency.

\item A Structural Key View (SKV) selection method is designed for each CR, such that the richest LF spatial information could be preserved with the minimum bit size and maximum SKV re-usability.

\item Disparity guided sparse coding over a learned perspective-shifted light field dictionary is proposed for the reconstruction of the entire LF based on the selected SKVs. The disparity guidance is vital in ensuring the quality and efficiency of the reconstruction.

\item By keeping the sparse coding approximation identical between encoder and decoder, only the residuals of non-key views, disparity map and the SKVs need to be compressed into the bit stream. Since the sparse coding coefficients are not to be transmitted, a high approximation quality could be achieved without the concern for large patch numbers (overlapping patch decomposition could be adopted) or too many coding coefficients for each patch.

\item The proposed codec is similar to a LF compressed sensing framework which can reconstruct the light field better with much fewer measurements (or number of sparse set examples in the sense of \cite{li2016scalable}). And because of the advantage from compressed sensing, we see that the reconstruction shows huge advantage especially in low bit rate cases.

\item We adopt the pseudo sequence method \cite{liu2016pseudo} for the compression of sparse coding residuals. This shows the flexibility of the proposed framework in the sense that it can be combined with existing compression methods. Since the sparse reconstruction takes zero bit-rate, a variety of pseudo sequence based LF codecs can be used for the compression of the reconstruction residuals.
\end{enumerate}

Experiments and comparisons have been carried out over the LF dataset \cite{2016LFdataset} which contains 12 benchmark LF images. The results show that the proposed \textit{SC-SKV} codec produces convincing compression results in terms of rate-distortion performance and visual quality compared with \textit{High Efficiency Video Coding} (\textit{HEVC}): with 47.87\% BD-rate reduction and 1.59 dB BD-PSNR improvement achieved on average, especially with up to 4 dB improvement for low bit rate scenarios.
 
The rest of the paper is organized as follows: Sec. \ref{sec_relatedWork} introduces related work. Sec. \ref{sec_system_intro} gives an brief overview of the LF data format and the proposed \textit{SC-SKV} codec. Sec. \ref{sec_codec_details} gives detailed descriptions of the codec components: in Sec. \ref{sec_structure} the formation procedures of the Light Field Dictionary (LFD) $D$ is reviewed; in Sec. \ref{sec_CR} the Coding Region (CR) segmentation and Structural Key View (SKV) selection is introduced; in Sec. \ref{sec_function_disparityMap} the disparity estimation process is explained; in Sec. \ref{sec_function_coding}, the disparity guided sparse coding over LFD based on SKVs is described; Sec. \ref{sec_function_residual} explains the residuals coding process. In Sec. \ref{sec_experiments} the evaluation results of the proposed \textit{SC-SKV} codec is presented. Finally, Sec. \ref{sec_conclusion} concludes the paper.

Throughout this paper, scalars are denoted by italic lower-case letters, vectors by bold lower-case letters, and matrices by upper-case ones, respectively. $\{\cdot\}^T$ denotes matrix operation of transpose. $vec\{\cdot\}$ denotes the operation of vectorizing a matrix into vector ($\mathbb{R}^{m\times n}\rightarrow\mathbb{R}^{(m\cdot n)\times 1}$).

\section{Related Work}\label{sec_relatedWork}

Conti et al. \cite{conti2016comm,conti2016hevc} proposed the concept of self-similarity compensated prediction, and Monteiro et al. \cite{Monteiro2016light} proposed to add locally linear embedding-based prediction to explore the inherent correlation within the MLIs. Field-of-View scalability could be achieved in these frameworks \cite{conti2013inter}. Li et al. \cite{li2016compression} proposed a bi-directional inter-frame prediction into the conventional \textit{HEVC intra} prediction framework for better compression results.

In \cite{magnor2000data} the authors modified the video compression techniques and proposed disparity-compensated image prediction among the LF views. Kundu et al. \cite{sakamoto2012study} exploited inter-view correlation by applying homography and 2D warping for view prediction. 
Relevant research in multi-view video coding \cite{shi2011efficient}, full parallax 3D video content compression \cite{Dricot2015full}, and holoscopic video coding \cite{conti2013inter} can be efficiently applied on LF data. Sakamoto et al. \cite{sakamoto2012novel} proposed to reconstruct the LF from multi-focus images, which contain mostly low frequency components and therefore easier to compress. Gehrig et al. \cite{gehrig2007distributed} studied LF compression through distributed coding. Principle Component Analysis (PCA) was utilized by Lelescu et al. \cite{lelescu2004representation} for de-correlating light field data.

Recently, pseudo sequence based methods achieved remarkable compression performance. One major variable under manipulation is the ordering of SVIs. And several options have been investigated such as the rotation order \cite{dai2015lenselet}, raster scan order \cite{perra2016high}, the hybrid horizontal zig-zag and U-shape \cite{zhao2016light} scan order, etc. Liu et al. \cite{liu2016pseudo} proposed to compress the center view as I-frame; the remaining views were set as P- or B-frames in a symmetric 2D hierarchical structure. Each view was assigned a layer. Views at higher level layer were coded after views at lower level layer and thus could be predicted from the latter. Liu's method showed significant compression improvement over the other methods in this category.

\begin{figure*}[!t]
	\centerline{\subfloat{\includegraphics[width=6.8in]{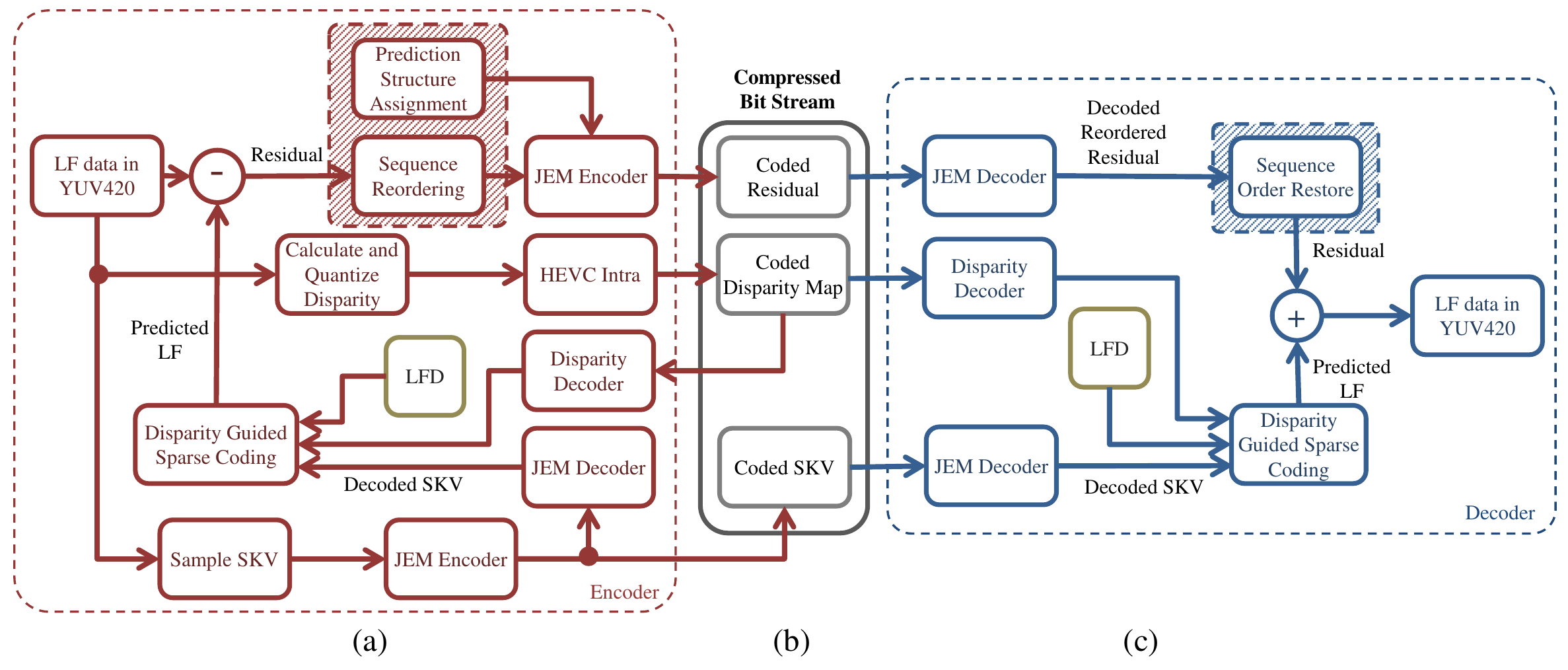}}}
	\caption{Block diagram of the proposed \textit{SC-SKV} codec: (a) The encoder; (b) Compressed bit stream; (c) The decoder. The shaded area boxes in (a) and (b) are implemented according to Li et al. \cite{liu2016pseudo}.}
	\label{fig_systemDrawing}
\end{figure*}

\section{Overview of the Proposed Light Field Codec}\label{sec_system_intro}


The proposed \textit{SC-SKV} codec focuses on processing LF images taken by Lytro Illum Camera, which is currently the mainstream LF camera on the market. The raw LF lenselet image from the camera can be decoded into a 2D array of LF sub-view images using the Matlab Light Field Toolbox \cite{dansereau2013decoding}. Fig. \ref{fig_viewSegment}(a) shows one decoded LF with angular dimension 15$\times$15. The proposed \textit{SC-SKV} codec is designed specifically for this angular configuration, although the coding region (CR) segmentation methodology can be flexibly extended to LFs of any angular dimensions.

A complete system diagram of the proposed \textit{SC-SKV} codec is shown in Fig. \ref{fig_systemDrawing}, and a brief introduction to the codec architecture will be given in this section. The implementation details will be explained later in Sec. \ref{sec_codec_details}.

\subsection{The Encoder}

As illustrated in Fig. \ref{fig_systemDrawing}, the proposed \textit{SC-SKV} codec compresses the LF data in three steps. 

First, a group of Structural Key Views (SKV) that best represent the spatial information of the LF are chosen and compressed with JEM codec. 

Secondly, the entire LF is decomposed into several Coding Regions (CR), and each CR is independently approximated with disparity guided sparse coding over a perspective shifted light field dictionary. The sparse coding is only based on the SKVs in the current CR. The sparse coding specifics (e.g. sparse coding coefficient number, target coding error, etc.) between the encoder and decoder are identical; therefore only the residuals of non-key views, the disparity map, and the SKVs need to be saved into the bit stream, while the sparse coding coefficients will be discarded since they can be re-estimated again in the decoder. The LF disparity map will be calculated and used in the sparse coding process, it will also saved into the bit stream with \textit{HEVC Intra}.

Finally, the SVI residual sequence from the approximation in the previous step is reordered and then each assigned a prediction parameter (such as P- or B-frame, hierarchy layer, QP value etc.), according to \cite{liu2016pseudo}. The sequence will be compressed with \textit{JEM} software \cite{jem2017jem}. Contents of the final compressed bit stream are shown in Fig. \ref{fig_systemDrawing}(b).

\subsection{The Decoder}

The decoder first uses \textit{JEM} to decode both the SKVs and the disparity map.

Secondly, disparity guided sparse coding is implemented based on the decoded SKVs. Identical sparse coding specifics (e.g. sparse coding coefficient number, target coding error, etc.) will be set here as in the encoder to ensure identical approximation for each CR.

Lastly, the LF residuals will be decoded by \textit{JEM}. Sequence order will be restored, and then added back to the sparse coding output.

The key procedures of the proposed codec, i.e., LF dictionary formation, coding region segmentation, SKV selection, disparity calculation, sparse coding, and residual coding will be explained in detail in Sec. \ref{sec_structure}. 

\section{The \textit{SC-SKV} Codec Details} \label{sec_codec_details}

\subsection{Light Field Dictionary via Perspective Shifting}\label{sec_structure}

To reconstruct LF based on selected SKVs, a light field dictionary (LFD) created via central view atom perspective shifting will be adopted for sparse coding \cite{chen2015light} \cite{chen2014light}. The LFD will be used as a global dictionary for all LF images. A brief review of the LFD detail will be given in this sub-section.

Discretized light field data are usually comprised of an array of images that were virtually taken from a closely positioned multi-camera array. The strictly equidistant viewing positions determine the linear configuration of disparity values among different SVIs. For a $8\times8$ discretized light field, the horizontal and vertical disparity ratios between each off-center views with respect to the center view position are shown in matrices $H$ and $V$ in Eqn. (\ref{eqn_latticeV}) respectively, where each matrix element $H(v),~V(v)$ represents the relative position of a LF view, $v$ is the respective view index.
\begin{equation}\label{eqn_latticeH}
H  = \left[{\begin{array}{*{20}c}
	{-3.5} &...&{-3.5} &{-3.5} &{-3.5} &... &{-3.5}\\
	{-2.5} &...&{-2.5} &{-2.5} &{-2.5} &... &{-2.5}\\
	{-1.5} &...&{-1.5} &{-1.5} &{-1.5} &... &{-1.5}\\
	{-0.5} &...&{-0.5} &{-0.5} &{-0.5} &... &{-0.5}\\
	{0.5} &...&{0.5} &{0.5} &{0.5} &... &{0.5}\\
	{1.5} &...&{1.5} &{1.5} &{1.5} &... &{1.5}\\
	{2.5} &...&{2.5} &{2.5} &{2.5} &... &{2.5}\\
	{3.5} &...&{3.5} &{3.5} &{3.5} &... &{3.5}\\
	\end{array}} \right],\notag
\end{equation}
\begin{equation}\label{eqn_latticeV}
V  = \left[{\begin{array}{*{20}c}
	{-3.5} &{-2.5} &{...} &{-0.5} &0.5 &{...} &2.5 &3.5\\
	{-3.5} &{-2.5} &{...}&{-0.5} &0.5 &{...} &2.5 &3.5\\
	: &: & &: &: & &:&:\\
	{-3.5} &{-2.5} &{...}&{-0.5} &0.5 &{...} &2.5 &3.5\\
	{-3.5} &{-2.5} &{...}&{-0.5} &0.5 &{...} &2.5 &3.5\\
	\end{array}} \right].
\end{equation}

Let's denote a scene point's \textbf{Unit disparity} as $dp$, which we define as the displacement (in unit of pixels) between adjacent views. $dp$ is a value only related to the scene point's distance to the camera. The actual disparity value between the sub-view $v$ and the central view is calculated as:

\begin{equation}\label{eqn_shearVector}
\mathbf{p}(dp,v)=dp\times[H(v),V(v)],~v=1,2,...,v_n,
\end{equation}
where $\mathbf{p}(dp,v)$ is a vector that indicates the disparities in horizontal and vertical directions respectively, and $v_n$ is the total number of SVIs.

Suppose we have a 2D image dictionary $G\in \mathbb{R}^{n_c\times k_c}$ for image patches of size $\sqrt{n_c}\times\sqrt{n_c}$, trained with a set of natural images \cite{Aharon2006}. Here $k_c$ is the number of dictionary atoms. Each column of $G$ is a dictionary atom, which is a square image patch $I_k\in\mathbb{R}^{\sqrt{n_c}\times\sqrt{n_c}}$ in its vectorized form $vec\{I_k\}\in\mathbb{R}^{n_c\times 1}$.

If we assume that pixels in a small image patch area around that scene point have the same Unit disparity $dp$, then we can predict the patch appearance in any sub-views via domain transformations. More precisely, if $I_{k,v}$ represents the off-center view $v$ transformed via the center view $I_k$, there exists an affine transformation $\tau_\mathbf{p}:~\mathbb{R}^2\rightarrow\mathbb{R}^2$, such that：
\begin{equation}\label{eqn_geometry}
I_{k,v}(x,y)=(\tau_\mathbf{p}\circ I_k)(x,y)=I_k(\tau_\mathbf{p}(x,y)).
\end{equation}

The transform operator $\tau$ takes in the parameter $\mathbf{p}$, which specifies the direction and magnitude of displacement for a specific view $v$. According to Eqn. \ref{eqn_shearVector}, once a scene point's Unit disparity $dp$ is given, the set of affine transforms $\{\tau_\mathbf{p(dp,1)},\tau_\mathbf{p(dp,2)},...,\tau_\mathbf{p(dp,v_n)}\}$ will be uniquely determined for all light field views. We define $\kappa_{dp}:~\mathbb{R}^{\sqrt{n_c}\times\sqrt{n_c}}\rightarrow\mathbb{R}^{(n_c\cdot v_n)\times1}$ as the operation that creates a virtual light field based on a 2D image patch $I$. $\kappa_{dp}$ includes a series of transformations on $I$ for each sub-view, and a final step of vectorization and concatenation:
\begin{align}
\kappa_{dp}\circ I= [vec\{\tau_{\mathbf{p}(dp,1)}\circ I\}^T, & vec\{\tau_{\mathbf{p}(dp,2)}\circ I\}^T,...,\\ \notag
& vec\{\tau_{\mathbf{p}(dp,v_n)}\circ I\}^T]^T.
\end{align}

Therefore, the output of $\kappa_{dp}\circ I_k$ will be a LF atom with Unit disparity $dp$. Following the method in \cite{chen2015light}, we group all light field atoms transformed with identical $dp$ as one dictionary segment $D_{dp}$:
\begin{equation}
D_{dp}= [\kappa_{dp}\circ I_1,\kappa_{dp}\circ I_2,...,\kappa_{dp}\circ I_{k_c}]\in\mathbb{R}^{(n_c\cdot v_n)\times k_c}.
\end{equation}
The final LFD $D$ combines all segments $D_{dp}$ that cover the entire predefined disparity space: $\{dp_1,dp_2,...,dp_{s_n}\}$:
\begin{equation}
D= [D_1,D_2,...D_{s_n}]\in\mathbb{R}^{(n_c\cdot v_n)\times(k_c\cdot s_n)}.
\end{equation}
Here $s_n$ denotes the defined number of unique disparity values. The strict linear configuration of parallax among different SVIs, and the grouping of atoms based on their Unit disparity will contribute to the efficiency of light field reconstruction used in our proposed \textit{SC-SKV} codec. For other implementation details of the LFD, such as the handling of boundary pixels, sub-pixel interpolation, please refer to our previous works in \cite{chen2014light} \cite{chen2015light}. Also note that although we assume unit disparity for each dictionary atom, when the dictionary coding is carried out on a overlapping patch basis, the disparity for the reconstructed LF is still considered pixel-wise.

\subsection{Coding Region Segmentation} \label{sec_CR}

\begin{figure}[!t]
	\centering
	\includegraphics[width=3.5in]{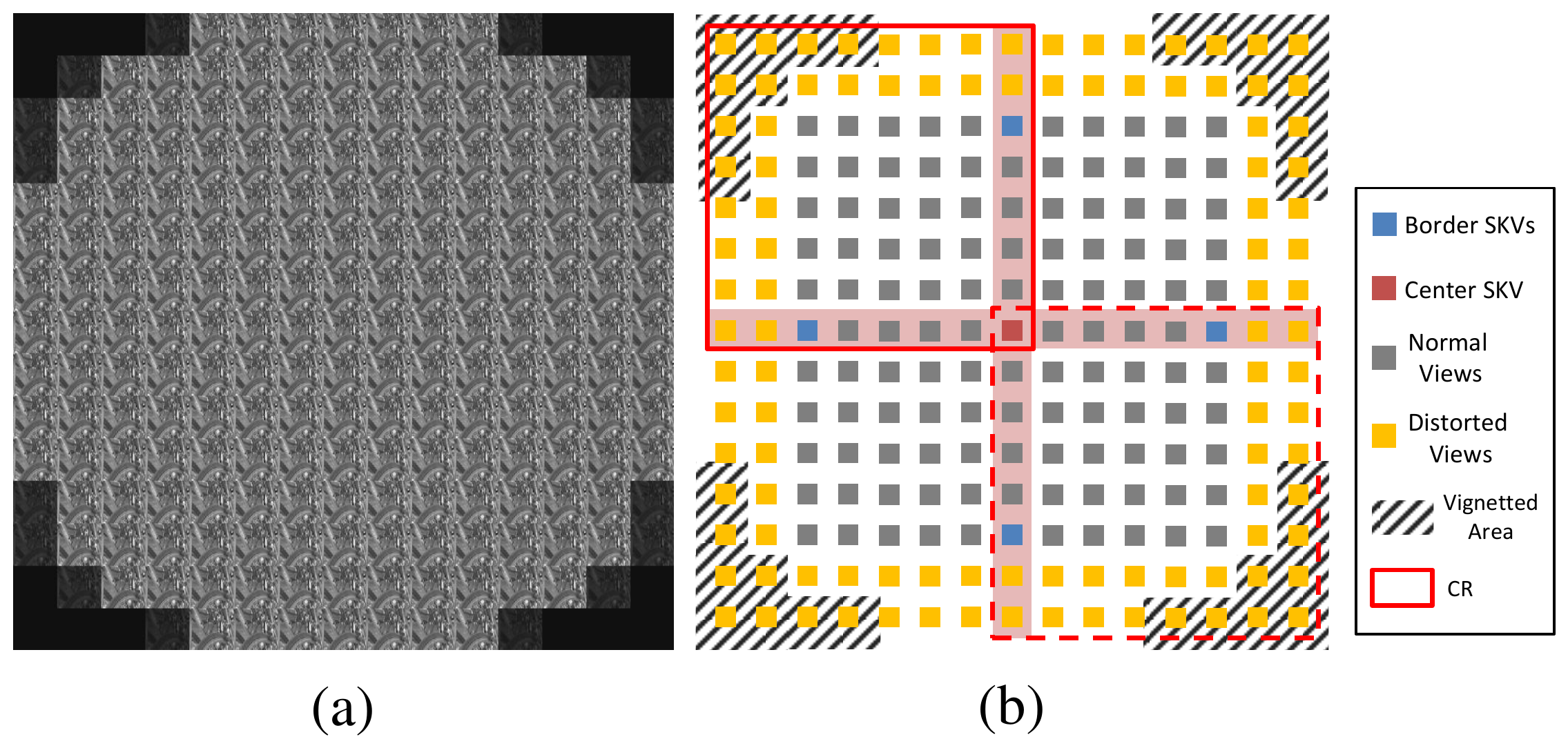}
	\caption{(a) An actual 15$\times$15 LF data converted with \cite{dansereau2013decoding} where obvious vignetting could be observed for the border views; (b) The large LF with 15$\times$15 sub-views is divided into 4 coding regions (CR) (8$\times$8).}
	\label{fig_viewSegment}
\end{figure}

Based on the $D$, LF of the same angular dimension ($v_n$=8$\times$8) can be easily represented. However for LFs of other angular dimensions, the dictionary need to be either trimmed or cascaded. When the angular dimension of LF increases, the data size of each LF image will increase significantly; the sparse coding for such data will simultaneously become much more complex and time-consuming. 


Our solution to avoid this problem is to segment the whole LF in the angular dimension into several coding regions (CR), each with a fraction of the original sub-view number. In this work, we aim at compressing the LF images from the LF dataset \cite{2016LFdataset}, which consists of 12 LF images with angular dimensions of 15$\times$15. A 15$\times$15 LFD is obviously too large to be manipulated; therefore we divided it into four 8$\times$8 CRs. As shown in Fig. \ref{fig_viewSegment}(b), the four CRs jointly cover the entire 15$\times$15 space, with only one line of border views overlapped. The overlapped views will facilitate structural key view re-use (to be explained in next sub-section). The angular dimensionality of each CR is identical to that of $D$. 

The concept of LF CR segmentation could be easily extended to other LF angular dimensions. When LF dimensions larger than 15$\times$15 is encountered, the number and location arrangement of CRs as well as the angular dimension for each CR could all be adaptively changed for efficient coverage of the entire LF.

\subsection{Structural Key View Selection}\label{sec_SKV}

As shown in Fig. \ref{fig_viewSegment}(b), 5 Structural Key Views (SKV) are selected by \textit{SC-SKV} codec for the reconstruction of the entire LF: one central view (shown in red), and four off-center views (shown in blue) in the overlapped CR borders. The SKVs are chosen under consideration of maximum re-usability amongst different CRs. As shown in \ref{fig_viewSegment}(b), the blue SKVs are shared between two neighboring CRs, and the red SKV is shared among all four CRs. 

The second consideration for SKV selection is the  preservation of LF spatial information. For each CR, three SKVs are involved for reconstruction (one red and two blues); and they are at or near the borders of each CR; therefore maximum disparity for each scene point could be observed, and this could be very beneficial during LF reconstruction.

The third consideration is the sub-view vignetting. As can be seen in \ref{fig_viewSegment}(a), the vignetting is most obvious at border views of the LF, especially on the four corners. The sub-views with obvious vignetting has been shaded in gray bars in \ref{fig_viewSegment}(b), and these views, although have richer spatial information, are not suitable to be chosen as SKVs. Besides the vignetting, optical aberration and distortion is another important factor. The sub-views with yellow colors are ones that suffer the most from it, and the image quality of these views are generally much lower than the central ones with obvious out-of-focus effects.

All of the above considerations validate our selection of SKVs for the LF reconstruction. 

\subsection{Disparity Map Estimation}\label{sec_function_disparityMap}

To take advantage of the LFD $D$, which has distinctive disparity segments $[D_1 ,D_2 ,...,D_{s_n} ]$, the disparity of the scene needs to be calculated first.

We adopt the voting method introduced in \cite{Kim2013}, which calculates a probability distribution for the entire disparity range. The probability distribution function value for a certain disparity $dp$ is quantized according to the total intensity differences between the LF views and their respective approximations by shifting the central view according to Eqn. (\ref{eqn_shearVector}) at $dp$.

The final output disparity map $P$ is determined to be disparity values with largest posterior possibilities. Along with $P$, an edge confidence $C_e$ can also be calculated by measuring the EPI intensity variance.  $C_e$ can be used to represent the confidence and correctness of the disparity map.

\subsection{Disparity Guided Light Field Sparse Coding}\label{sec_function_coding}

Based on the SKVs, we now explain how to reconstruct the entire LF. Let's denote the segmented LF data in the current CR as $L_R(c)$ ($c$ is the CR index).
Let $\mathbf{l}_{i,v}\in\mathbb{R}^{n_v\times1}$ denote a vectorized patch at location $i$ on sub-view $v$, and let $\mathbf{l_{i}}\in\mathbb{R}^{(n_v\cdot v_n)\times1}$ denote the concatenation of all vectorized patches at location $i$ from all the sub-views in the current CR:
\begin{equation}
\mathbf{l_{i}}= [\mathbf{l}^T_{i,1},\mathbf{l}^T_{i,2},...,\mathbf{l}^T_{i,v_n}]^T,
\end{equation}

As has been explained in Sec \ref{sec_SKV}, the 15$\times$15 LF is divided into 4 CRs each of dimension 8$\times$8. For each CR, 3 SKVs are involved. Let $\mathbf{k}_{i}\in\mathbb{R}^{3\cdot n_v\times1}$ denote the concatenation of patch $i$ from all SKVs in the current CR:
\begin{equation}
\mathbf{k}_{i}= [\mathbf{l}^T_{i,v_1},\mathbf{l}^T_{i,v_2},\mathbf{l}^T_{i,v_2}]^T,~(v_1,v_2,v_3\in \Omega_c),
\end{equation}
where $\Omega_c$ denotes the set of SKV indices for the current CR $c$. We have:
\begin{equation}
\mathbf{k}_{i}=\Phi_c \mathbf{l}_{i}.
\end{equation}
Here $\Phi_c$ is a linear operation matrix that extracts the current CR's SKVs from the full vectorized LF $\mathbf{l}_{i}$.

Since the LF data have a strong spatio-angular correlation, they can be sparsely represented by a few signal examples \cite{Marwah2012} \cite{elad2012sparse}. With the perspective-shifted LF dictionary $D$, a light field signal $\mathbf{l}_{i}$ can be represented as:
\begin{equation}\label{eqn_sensingMatrix}
\mathbf{l}_{i}= D\mathbf{\alpha}_i.
\end{equation}
where $\mathbf{\alpha}_i\in\mathbb{R}^{(s_n\cdot k_c)\times 1}$ is the LF CR's representation coefficient vector over the LF dictionary $D$. And the set of SKVs for the CR can be expressed as:
\begin{equation}
\mathbf{k}_{i}=\Phi_c D \mathbf{\alpha}_i,
\end{equation}

As previously explained, the LF dictionary can be divided into multiple segments $D  = [D_1 ,D_2 ,...,D_{s_n} ]$: the atoms that have the same disparity value are grouped as one segment. The sparse coding coefficient $\mathbf{\alpha}_i$ can also be divided into multiple segments $\mathbf{\alpha}_i = [\mathbf{\alpha}_{i,1},\mathbf{\alpha}_{i,2},...,\mathbf{\alpha}_{i,s_n}]$; each corresponds to a dictionary segment, and only one disparity segment should be non-zero. 

Let's denote $\Theta(\cdot)$ as a linear operator that quantizes disparity values into integers that correspond to the disparity segment indices in $D$. For coefficient segments $\mathbf{\alpha}_{i,dp},~(dp= 1,2,...s_n)$ whose indices $dp$ are different from the calculated LF disparity: $dp \ne \Theta(P_i)$, $\mathbf{\alpha}_{i,dp}=0$; and for the segment $dp= \Theta(P_i)$, the following optimization problem is to be solved:
\begin{equation}\label{eqn_greedyPursuit}
\begin{array}{l}
\min\left\| \mathbf{\alpha}_{i,\Theta (P_i)}\right\|_0,  \\ \\
s.t.~\left\| {\mathbf{k}_{i} - 
\Phi_c D \mathbf{\alpha}_{i,\Theta(P_i)}} \right\|_2 \le \varepsilon,
\end{array}
\end{equation}
where $\epsilon$ is the coding error threshold. 

The $l_0$-norm $||\cdot||_0$ in Eqn. (\ref{eqn_greedyPursuit}) calculates the number of non-zero elements in the vector $\mathbf{\alpha}_{i,\Theta (P_i)}$. To solve the problem, we adopt the greedy Orthogonal Matching Pursuit (OMP) algorithm \cite{Elad2006}. The OMP coding will be very computationally efficient since only a small segment of the dictionary is used for coding ($(n_v\cdot v_n)\times k_c$), instead of the entire dictionary, which has a much larger dimension of ($(n_v\cdot v_n)\times(s_n\cdot k_c)$). This leads to a much more efficient light field reconstruction algorithm both in terms of both computation complexity and reconstruction quality \cite{chen2015light}.

The coding coefficients $\{\mathbf{\alpha}_i\}$will be used to reconstruct each over-lapping patches, and the final $L_R(c)$ will be the average of all patches. After all CRs are coded separately, the entire LF will be concatenated.

\subsection{Residual Coding}\label{sec_function_residual}


The approximation process in the previous step efficiently removes the spatial and angular correlations among the SVIs. However some structural correlations in the LF is still unavoidably left in the residuals. To effiiently compress the approximation residuals,
we choose the codec introduced in \cite{liu2016pseudo}, which is modified based on the \textit{JEM} software \cite{jem2017jem} specifically designed for LF data. According to \cite{liu2016pseudo} the SVI residual sequence is first reordered and then each assigned a different compression parameter. The center view will be compressed as I-frame; the remaining views will be compressed as P- or B-frames in a symmetric 2D hierarchical structure. Each view is assigned a layer, with different QP values and different prediction relationships separately assigned for each layer. The sequence will be compressed with \textit{JEM}.


\section{Experiment Results and Discussion}\label{sec_experiments}

In this section, we carry out extensive experiments to evaluate the proposed \textit{SC-SKV} codec, and compare it with the state-of-the-art methods.

Following the steps introduced in Sec. \ref{sec_structure}, a perspective-shifted LF dictionary $D$ is prepared with the following parameters: the global image dictionary $G\in\mathbb{R}^{400\times400}$ is trained with a set of natural image patches
\footnote{\label{ft_trainingImgs}The 2D dictionary is trained with 10 benchmark natural images: lena, barbara, boat, peppers, etc.} \cite{Aharon2006}. For the LF dictionary $D$: $\sqrt{n_v}\times\sqrt{n_v}=8\times8$, and total sub-view number $v_n=8\times 8=64$. The discretized disparity range for $D$ is set as a 21$\times$1 vector: $[$-3.0, -2.7, -2.4, -2.1, -1.8, -1.5, -1.2, -0.9, -0.6, -0.3, 0, 0.3, 0.6, 0.9, 1.2, 1.5, 1.8, 2.1, 2.4, 2.7, 3.0$]$. Therefore, the final LFD is $D \in\mathbb{R}^{(64\cdot64)\times(21\cdot400)}=\mathbb{R}^{4096\times8400}$.

\begin{figure}[!t]
\centering
\includegraphics[width=3.5in]{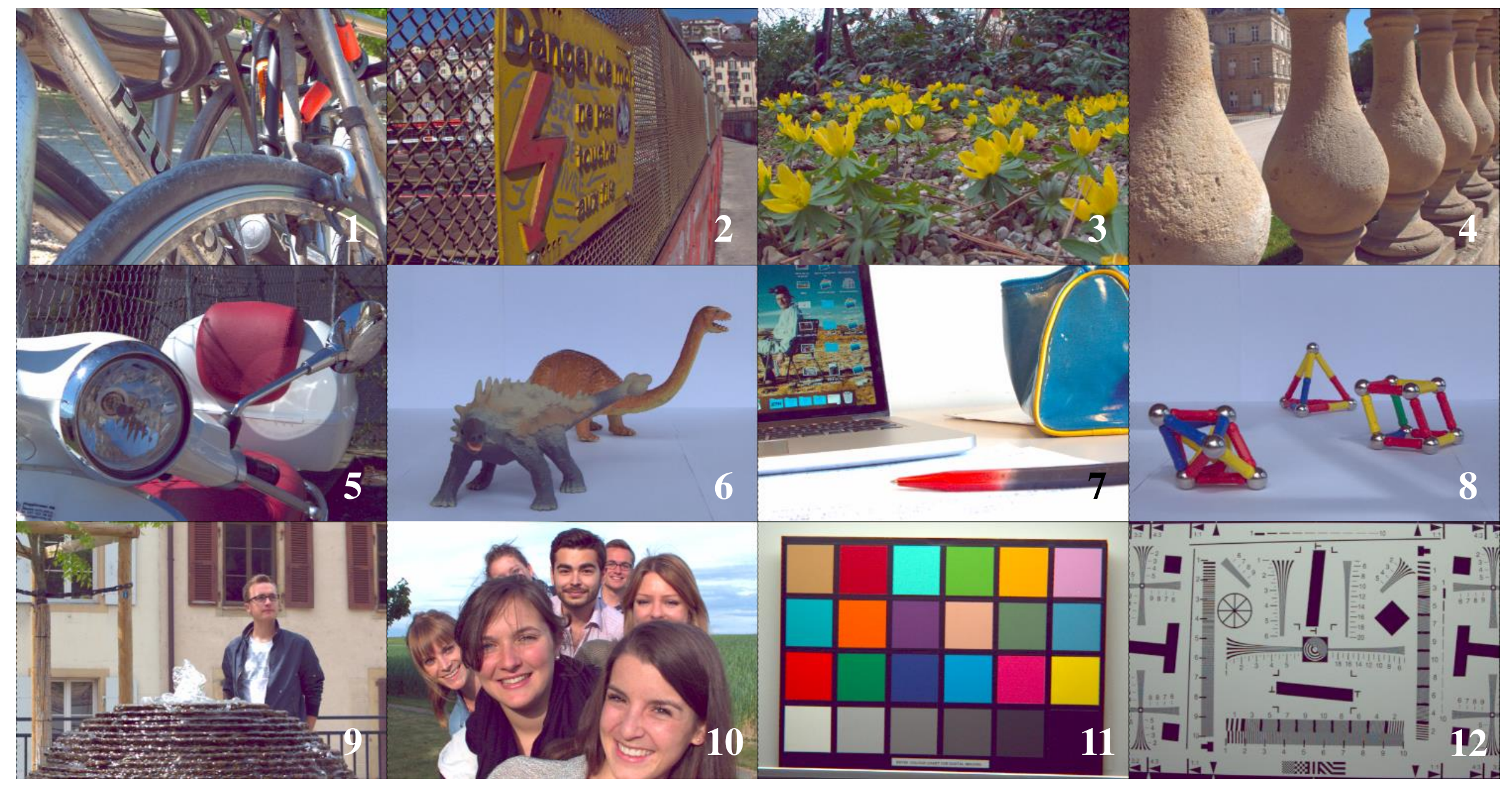}
\caption{Light field image data set thumbnails \cite{2016LFdataset}: \textit{I01 Bikes}, \textit{I02 anger De Mort}, \textit{I03 Flowers}, \textit{I04 Stone Pillars Outside}, \textit{I05 Vespa}, \textit{I06 Ankylosaurus \& Diplodocus 1}, \textit{I07 Desktop}, \textit{I08 Magnets 1}, \textit{I09 Fountain \& Vincent 2}, \textit{I10 Friends 1}, \textit{I11 Color Chart 1}, \textit{I12 ISO Chart 12} are arranged according to the order on the right.}
\label{fig_LFImageDataSet}
\end{figure}

The LF dataset from \cite{2016LFdataset} will be used for evaluation, which consists of 12 LF images taken by the Lytro camera at various natural and manually set-up scenes. The LF image thumbnails from the data set are shown in Fig. \ref{fig_LFImageDataSet}. Each LF image has been decoded from the original 12-bit lenselet images using Matlab Light Field Toolbox \cite{dansereau2013decoding} into a 3D LF structure of dimension $432\times 624\times 225$. The original RGB data have been downsampled to YUV 420 before we apply our codec. 

\begin{figure}[!t]
	\centering
	\includegraphics[width=2in]{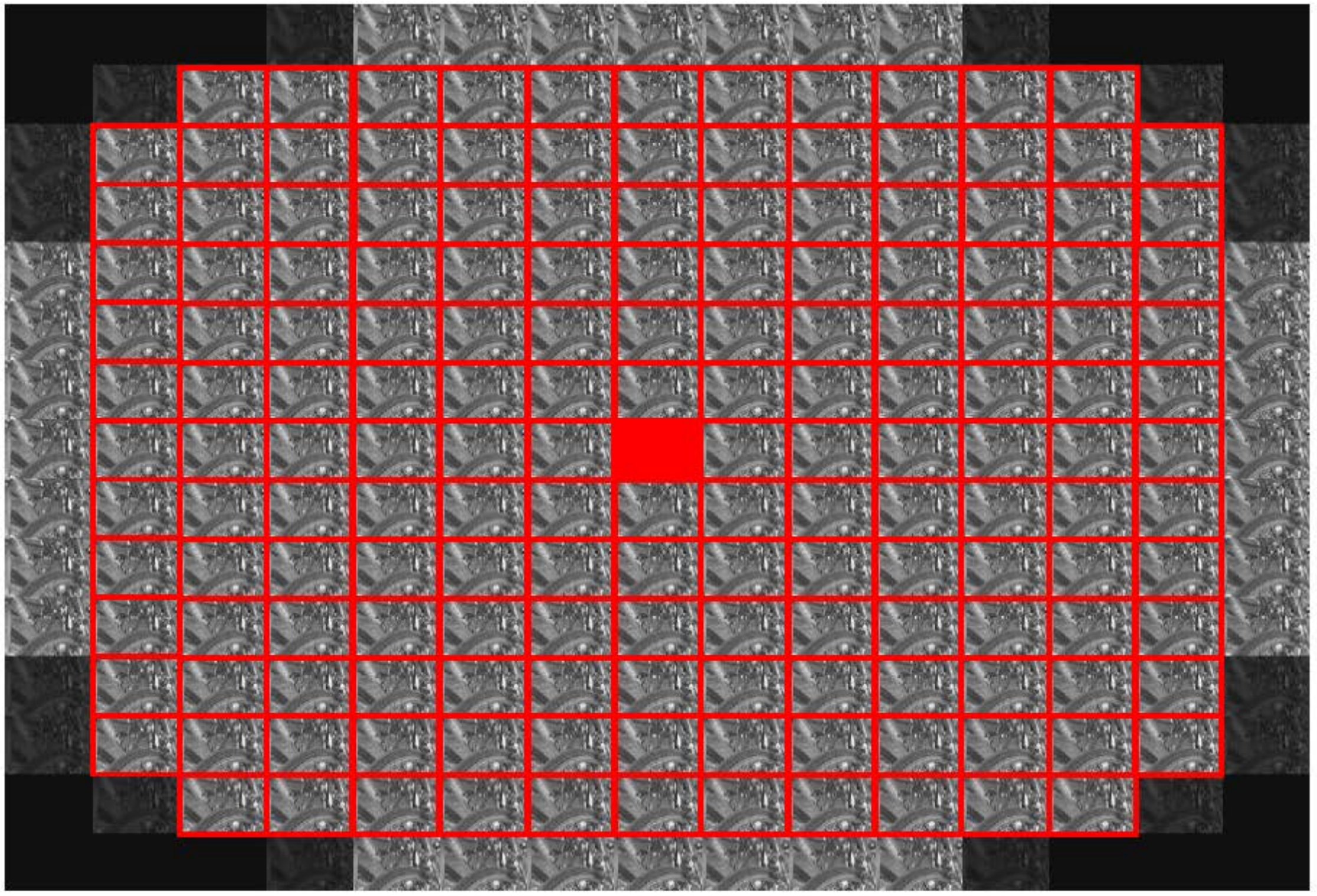}
	\caption{SVIs selected for evaluation. The vignetted border SVIs are excluded. The central view is highlighted in red.}
	\label{fig_fieldConfiguration}
\end{figure}

In order to accurately evaluate the LF reconstruction performance, the vignetted SVIs at the border of the LF will be excluded from our evaluation. As shown in Fig. \ref{fig_fieldConfiguration}, only the SVIs in red rectangles will be selected for all subsequent experiments. Consequently, a total  of 13$\times$13$-$4$=$165 SVIs will be included in the LF pseudo sequence.

\subsection{Compression of SKVs} \label{sec_exp_sampled}

The compression of SKVs is of vital importance for the reconstruction of each CR $L_R(c)$ and consequently, the entire LF. In our designed SKV setups introduced in Sec. \ref{sec_SKV}, 5 SKVs need to be compressed. These views are arranged as pseudo-sequence and compressed using either \textit{JEM} or \textit{HEVC} in ``Low-delay P-main" mode. 

\begin{figure}[!t]
	\centering
	\includegraphics[width=3.5in]{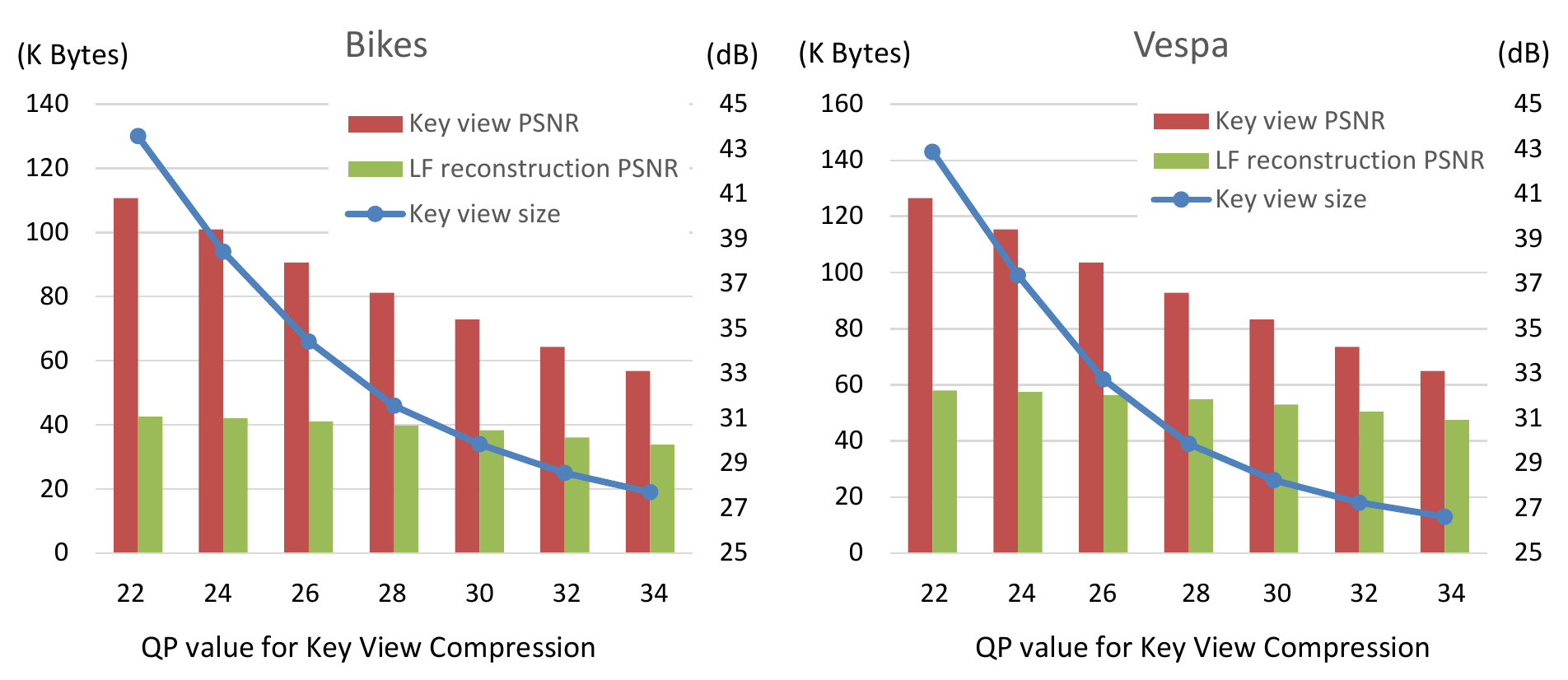}
	\caption{Compression of SKVs and its impact on LF reconstruction.}
	\label{fig_sampleSize}
\end{figure}

To determine a reasonable QP, we investigate the impact of the compression quality of SKVs on that of the final LF reconstruction, and different QPs are tested for two of the LF images \textit{I01\_Bikes} and \textit{I05\_Vespa}. The results are shown in Fig. \ref{fig_sampleSize}. 

As can be seen, as the QP increases from 22 to 34, both the bit size and the decoded PSNR for the SKVs decrease sharply. However, the overall LF reconstruction PSNR based on these SKVs decreases much slower. For both LF images in this experiment, we found that QP=30 is a good compromise between the SKV compression size and overall LF reconstruction quality. Therefore, QP=30 will be used for all other LF images in the dataset in the following experiments. More advance rate distortion optimization based bit allocation schemes can be developed to improve the overall performance \cite{hou2015compressing} \cite{hou2014highly}, which is left for our future work.

\begin{figure}[!t]
	\centering
	\includegraphics[width=2.4in]{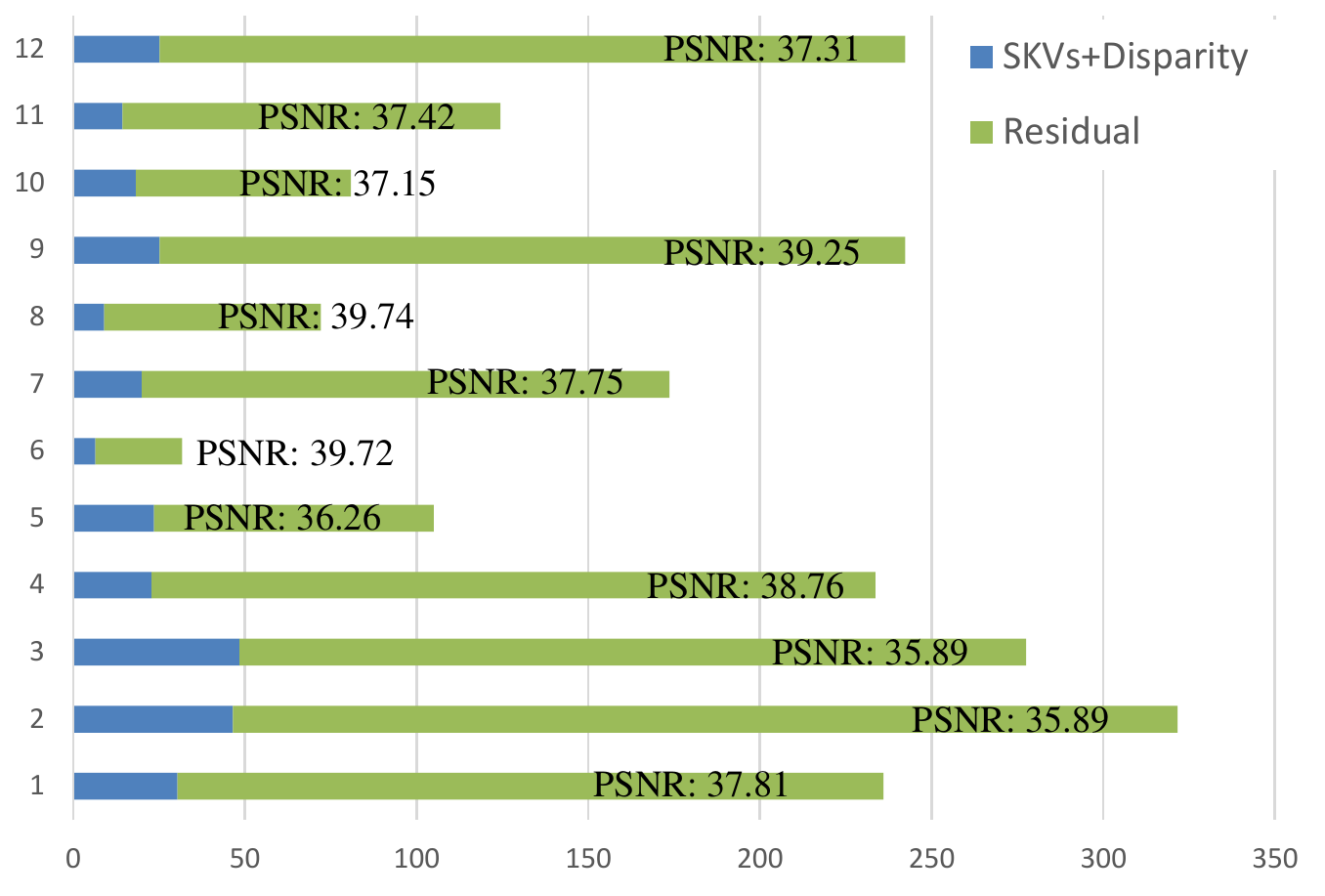}
	\caption{Bit portion allocated for the sampled key frames, disparity maps, and residuals for each LF data. Horizontal axis denotes total bit stream size (in KBytes), the vertical axis denotes image index in the LF dataset \cite{2016LFdataset}.}
	\label{fig_bitAllocation}
\end{figure}

With the QP value for the compression of SKVs and the QP value for the compression of LF residuals both fixed as 30, we evaluate the bit size portions of each components in the final compressed bit stream. As shown in Fig. \ref{fig_systemDrawing}(b), these components include: compressed SKVs, discretized disparity map, and the LF residuals. All components are necessary for the LF reconstruction in the decoder end. The bit size ratio for each components have been plotted in Fig. \ref{fig_bitAllocation} for each LF images in dataset \cite{2016LFdataset}, the LF reconstruction PSNR is also shown. As can be seen, the compressed SKVs and the disparity map take up less than 25\% of the entire bit stream. When the compression QP for the LF residuals decreases, even less bit portion will be taken up by the SKVs.

\subsection{Evaluation of the Disparity Guided Sparse Coding}

As a key component of the \textit{SC-SKV} codec, it is important to evaluate the quality of the LF approximation via disparity guided sparse coding, the output of which are expected to remove the LF spatial and angular correlations as much as possible. For all LF images, the maximum sparse coding coefficient number for each patch is limited at 30, and the target coding error is set as $\epsilon=5$ (out of image intensity range [0,255], defined in Eqn. \ref{eqn_greedyPursuit}).

\begin{figure}[!t]
\centering
\includegraphics[width=3.4in]{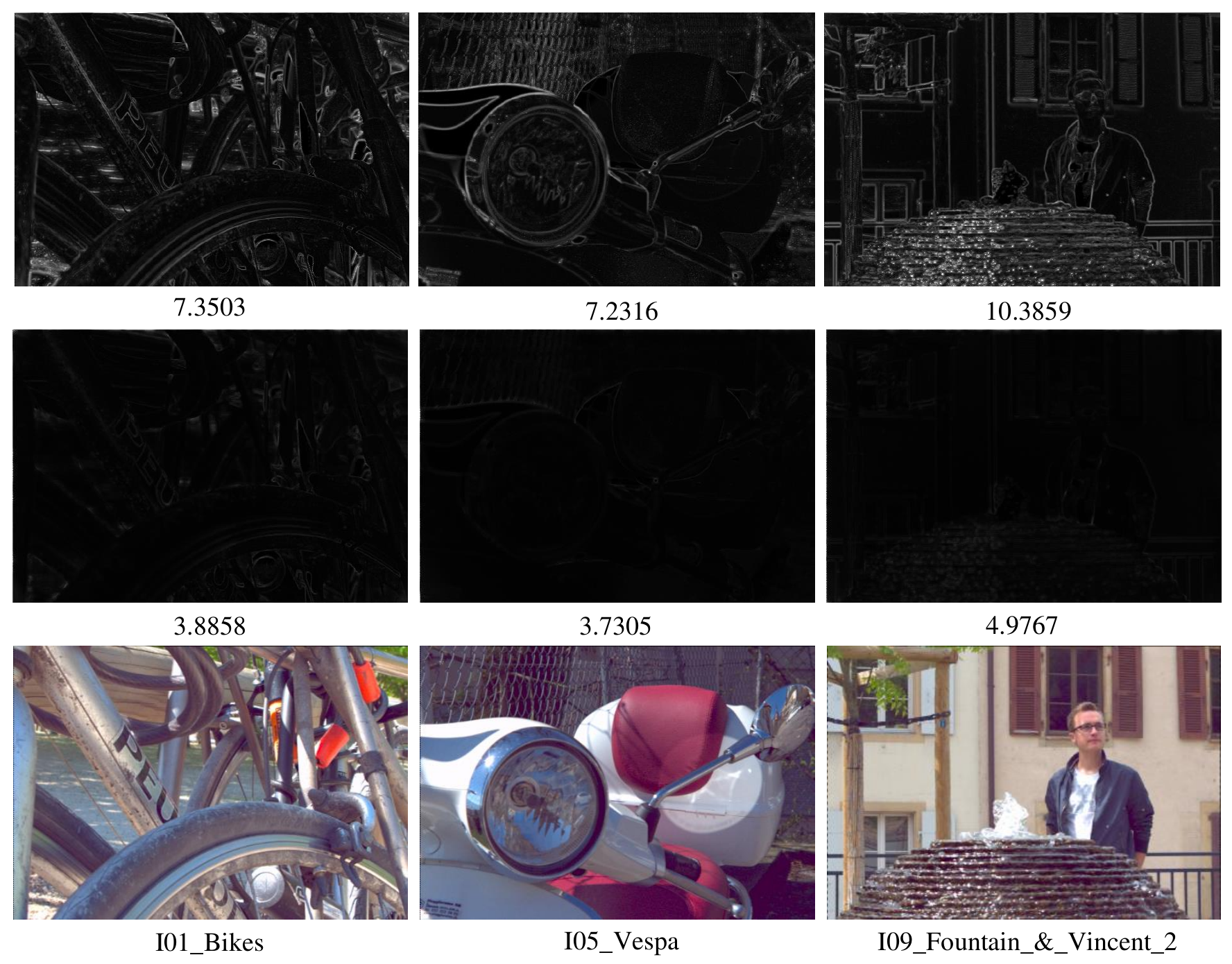}
\caption{Coding residuals for the LF \textit{Bikes}, \textit{Vespa}, and \textit{Fountain \& Vincent 2} in each columns respectively. The first row are the mean absolute differences of each view with the central view. The second row are the mean absolute residuals of all views after sparse coding approximation. The mean absolute values for each image are listed below.}
\label{fig_codingResidual}
\end{figure}

On the top row of Fig. \ref{fig_codingResidual}, the mean absolute differences between each sub-views and the center views are showns for three LF data: \textit{I01 Bikes}, \textit{I05 Vespa}, and \textit{I09 Fountain \& Vincent 2}. 
As a comparison, the second row in Fig. \ref{fig_codingResidual} shows the mean absolute residuals of all views after sparse coding reconstruction. As can be seen, for the second row, most of the inter-view spatial and angular correlations have been properly removed. The differences between the first two rows in Fig. \ref{fig_codingResidual} explains the difference in compression performances between the proposed \textit{SC-SKV} codec and the \textit{HEVC} codec. The detailed compression for each LF image in the dataset will be shown in Sec. \ref{sec_exp_comparison}.

\subsection{Comparison of Overall Rate-Distortion Performance} \label{sec_exp_comparison}

We evaluate the proposed \textit{SC-SKV} codec on the light field data set \cite{2016LFdataset}. Reconstruction PSNR are calculated between the decoded LF data, and the uncompressed LF data at different bit rates (measured in bit-per-pixel (bpp)). The PSNR for each Y, U, V chanel are calculated as:
\begin{equation}\label{eqn_channelPSNR}
\text{PSNR}= 10\log_{10}\frac{255^2}{\frac{1}{v_n}\sum_{v}[I_v-I'_v]^2},
\end{equation}
where $I_v$ stands for the original SVI $v$, and $I'_v$ stands for the reconstructed SVI $v$. The average PSNR of the Y, U, V channels are calculated according to Eqn. \ref{eqn_avgPSNR}:
\begin{equation}\label{eqn_avgPSNR} 
\text{PSNR}_\text{YUV}= \frac{6\times \text{PSNR}_\text{Y}+ \text{PSNR}_\text{U}+ \text{PSNR}_\text{V}}{8}.
\end{equation}


Three methods are chosen for performance evaluation:
\begin{enumerate}
\item \textit{HEVC-EQ}: The LF data is arranged as pseudo-sequence in column scan order. Only the first frame is set as I-frame; the rest are all set as P-frames. The \textit{HEVC} codec will be used in ``Low-delay P-Main'' mode to compress the entire sequence. The QP values are fixed equal for all SVIs.
\item \textit{Liu et al.}: Proposed in \cite{liu2016pseudo}, in which the LF SVI sequence is first reordered and then each assigned a different compression parameter. The center view will be compressed as I-frame, the remaining views will be compressed as P- or B-frames in a symmetric 2D hierarchical structure. Each view is assigned a layer, with different QP values and different prediction relationships separately assigned for each layer. The sequence is compressed with \textit{JEM} software \cite{jem2017jem}.
\item \textit{Proposed}: Our proposed codec, in which the SVIs will first be approximated via disparity guided sparse coding over LFD based on SKVs. The approximation residual sequence will be compressed using the method in \cite{liu2016pseudo}. 
\end{enumerate}
 
\begin{figure*}[!t]
\centerline{\subfloat{\includegraphics[width=7in]{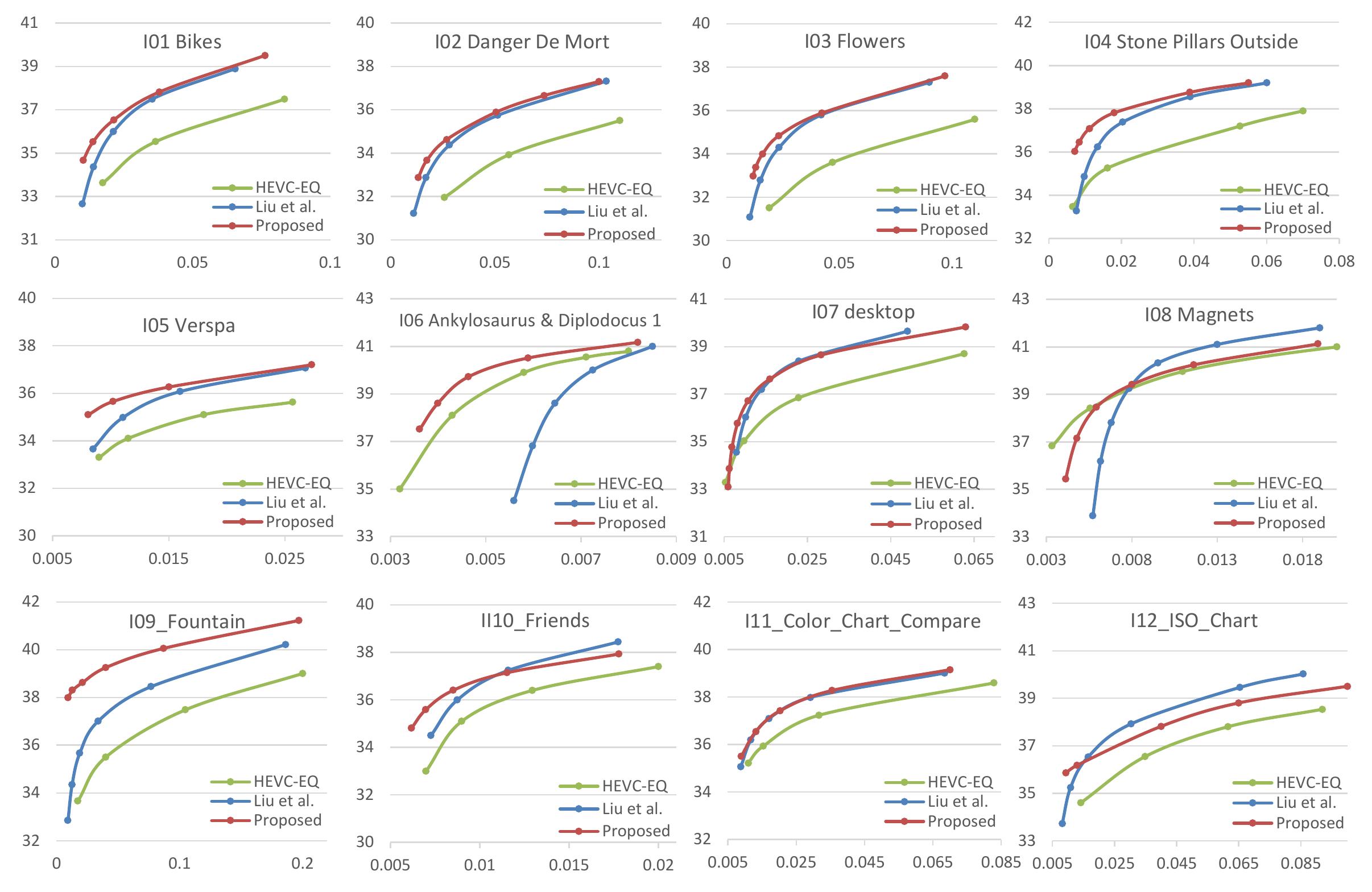}}}
\caption{LF reconstruction PSNR for LF dataset \cite{2016LFdataset}. The PSNR for YUV components average at different compression ratios are shown for the \textit{HEVC-EQ} codec, Liu et al.'s codec \cite{liu2016pseudo}, and our proposed \textit{SC-SKV} codec. The horizontal axis denotes the bit per pixel value (bpp), and the vertical axis denotes full-view reconstruction PSNR (dB).}
\label{fig_PSNR_dataset}
\end{figure*}

The rate-distortion curves for the three competing methods are shown in Fig. \ref{fig_PSNR_dataset}. The PSNR values shown on the figure are YUV channel average calculated according to Eqn. (\ref{eqn_channelPSNR}) and (\ref{eqn_avgPSNR}). As can be seen, the reconstruction PSNR of the proposed \textit{SC-SKV} codec is consistently 2 dB higher at all bit rates for most scenes as compared with \textit{HEVC-EQ}. For the exception of \textit{I06} and \textit{I08}, we can see the bpp range of the horizontal axis is much smaller (around 0.01) than all other LF images. This indicates that the compression performance from both \textit{HEVC-EQ} and \textit{Proposed} is already excellent even in extremely low bit-rates.
	
Still in Fig. \ref{fig_PSNR_dataset}, we can see that the proposed \textit{SC-SKV} codec shows better compression performance than \textit{Liu et al.}, especially under low bit rate scenarios. For example, \textit{SC-SKV} is 6dB better than \textit{Liu et al.} at bbp= 0.02 for \textit{I09}. \textit{SC-SKV} shows a consistent advantage over all bit rates than \textit{Liu et al.} in natural LF images (\textit{I01}, \textit{I02}, \textit{I03}, \textit{I04}, \textit{I05}, and \textit{I09}). For some manually set-up scenes (\textit{I06}, \textit{I07}, \textit{I11}, and \textit{I12}) the advantage of \textit{SC-SKV} codec is smaller but still obvious under low bit rate. For higher bit rate scenarios, \textit{Liu et al.} performs slightly better. From Fig. \ref{fig_LFImageDataSet}, it can be seen that these scenes are arranged similarly with small foreground objects in front of large uniform backgrounds. We believe for higher bit rate scenarios, if a more flexible QP value is set for the compression of SKVs, the performance will improve. A more advanced bit allocation optimization method can be designed for this scenario, which we leave for our future work.


\begin{table}
\centering
\caption{BD-PSNR/BD-BR: Comparison of proposed \textit{SC-SKV} codec with \textit{HEVC-EQ} and \textit{Liu}'s codec \cite{liu2016pseudo}.}
\begin{tabular}{ | >{\centering\arraybackslash}m{1cm} | >{\centering\arraybackslash}m{1.2cm} | >{\centering\arraybackslash}m{1.2cm} | >{\centering\arraybackslash}m{1.2cm} |>{\centering\arraybackslash}m{1.2cm} |}
\hline
\multirow{2}[4]{*}{No.}
&\multicolumn{2}{c|}{HEVC-EQ} &\multicolumn{2}{c|}{Liu et al.}
\\ \cline{2-5}
& BD-PSNR & BD-BR & BD-PSNR & BD-BR \\ \rowcolor{mygray} \hline \hline
I01 &+2.27	&-61.63\%	&+0.74	&-18.48\% \\  \hline
I02 &+2.32	&-64.77\%	&+0.39	&-13.78\% \\\rowcolor{mygray} \hline
I03 &+2.54	&-70.13\%	&+0.39	&-13.64\% \\ \hline
I04 &+2.31	&-75.98\%	&+1.04	&-33.59\% \\\rowcolor{mygray} \hline
I05	&+1.59	&-58.27\%	&+0.50	&-17.61\% \\ \hline
I06 &+0.73	&-13.93\%	&+2.01	&-33.74\% \\\rowcolor{mygray} \hline
I07 &+0.72	&-26.83\%	&-0.40	&+15.72\% \\ \hline
I08 &-0.13	&+2.87\%	&+0.06	&-11.49\% \\\rowcolor{mygray} \hline
I09 &+3.41	&-89.54\%	&+2.21	&-68.45\% \\ \hline
I10 &+1.28	&-32.49\%	&+0.08	&-6.26\% \\\rowcolor{mygray} \hline
I11 &+0.88	&-39.96\%	&+0.05	&-2.98\% \\ \hline
I12 &+1.11	&-43.81\%	&-0.28	&+24.86\%\\\rowcolor{mygray} \hline\hline
\textbf{Average} &\textbf{+1.59}	&\textbf{-47.87\%}	&\textbf{+0.57}	&\textbf{-14.95\%} \\ \hline
\end{tabular}
\label{tbl_BD_PSNR_RATE}
\end{table}

Table \ref{tbl_BD_PSNR_RATE} shows the BD-PSNR and BD-BR comparisons \cite{bjontegaard2001calcuation} of the proposed \textit{SC-SKV} codec with reference to \textit{HEVC-EQ} and \textit{Liu et al}, respectively. As can be seen, for \textit{I04}, the highest 89\% bit rate reduction and 3.41 dB PSNR increase could be achieved. An average 47.87\% bit rate reduction and 1.59 dB PSNR increase could be achieved for \textit{SC-SKV} as compared to \textit{HEVC-EQ}, and 14.95\%, 0.57 dB achieved as compared with \textit{Liu et al}.

\subsection{LF Reconstruction Visual Quality Demonstration}

\begin{figure}[!t]
	\centering
	\includegraphics[width=5.4in]{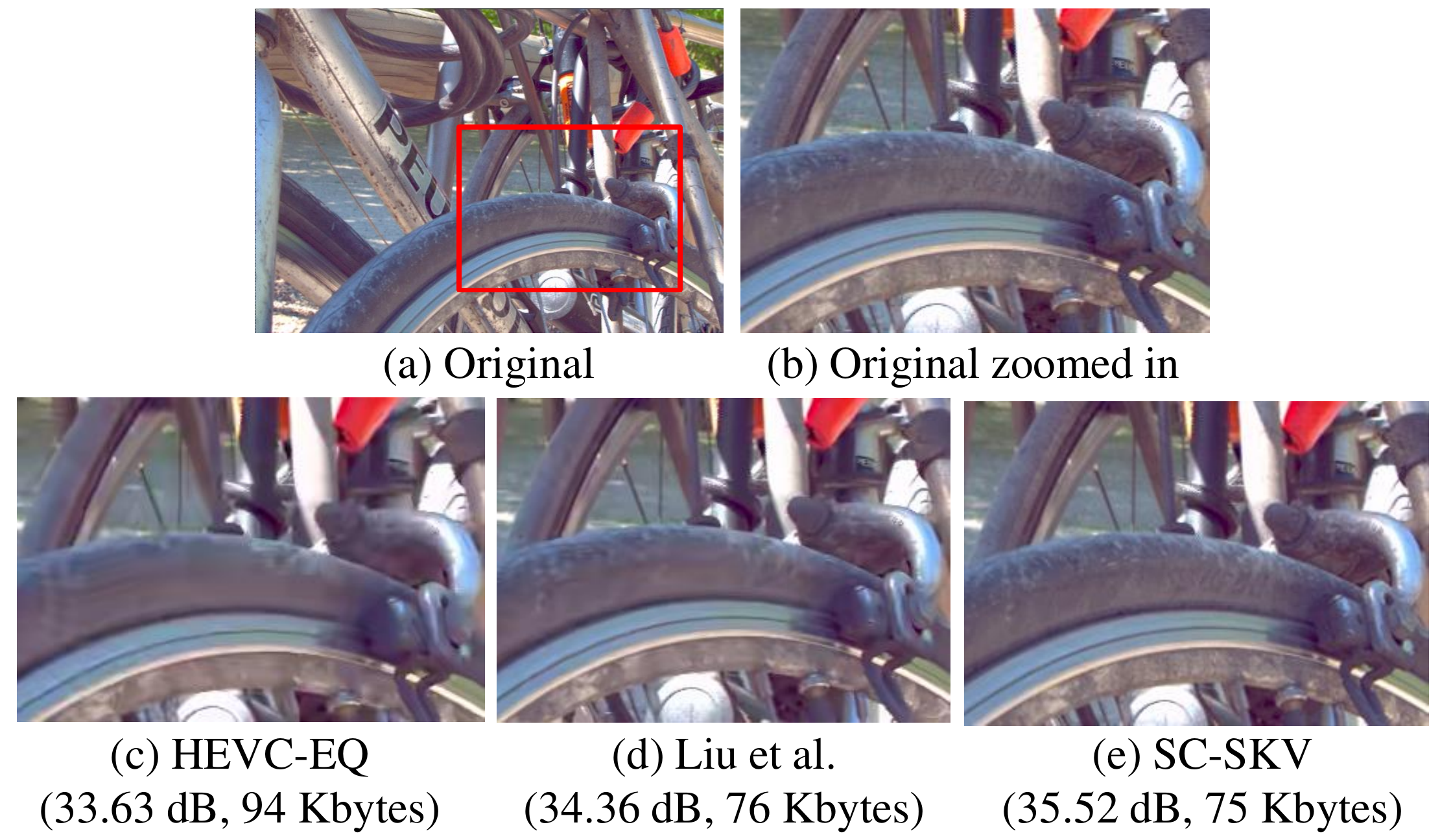}
	\caption{Visual comparison of decoded central view for the LF \textit{I01 Bikes} from different codecs. (b) is the original zoom-in on the red rectangle in (a). (c), (d), (e) are the zoom-in LF central views decoded from compressed bit stream of 94 KBytes by \textit{HEVC-EQ}, 76 KBytes by \textit{Liu et al.}, and 75 KBytes by \textit{SC-SKV}, respectively.}
	\label{fig_compareBikes}
\end{figure}

\begin{figure}[!t]
	\centering
	\includegraphics[width=5.4in]{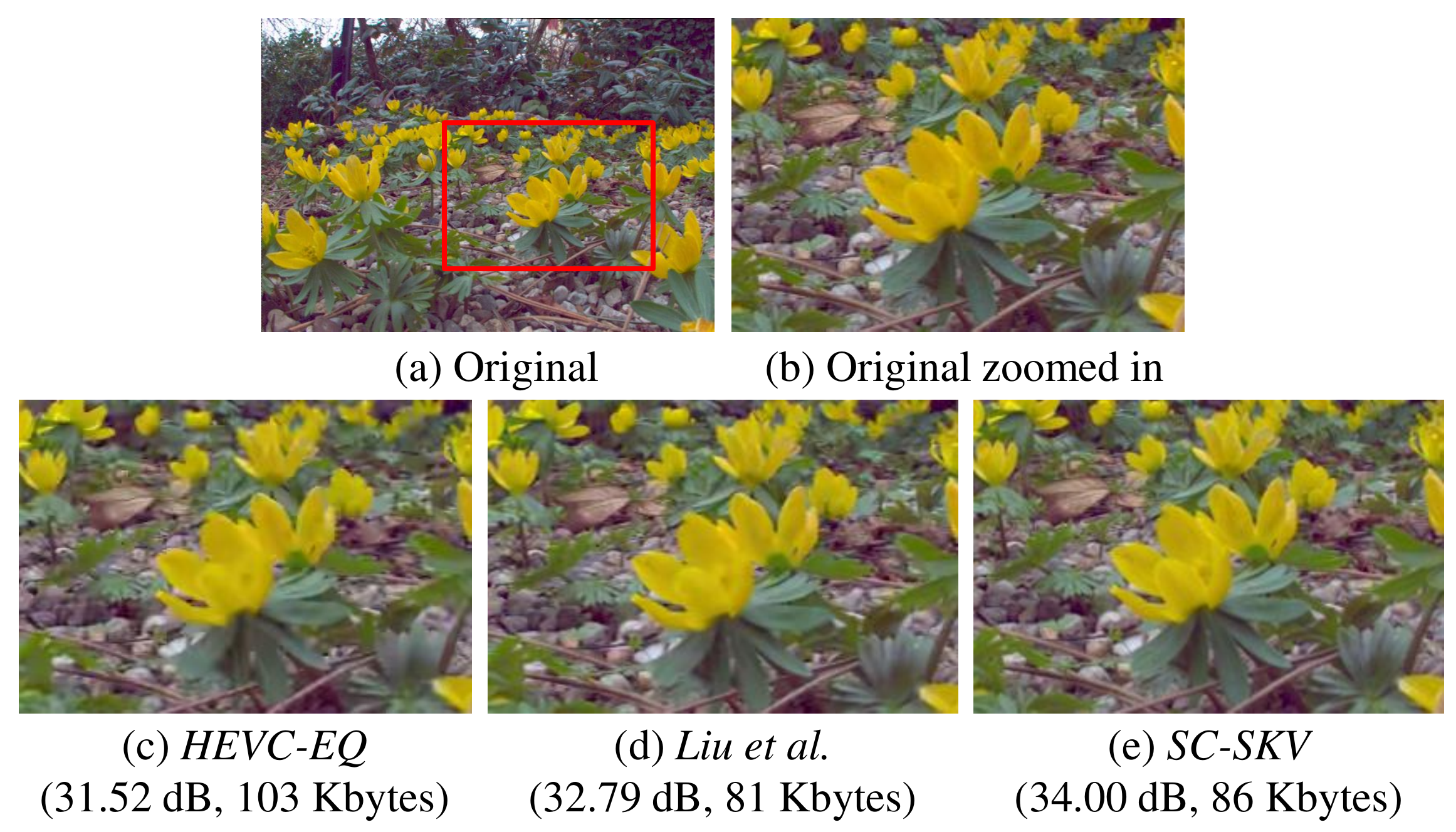}
	\caption{Visual comparison of decoded central view for the LF \textit{I03 Flowers} from different codecs. (b) is the original zoom-in on the red rectangle in (a). (c), (d), (e) are the zoom-in LF central views decoded from compressed bit stream of 103 KBytes by \textit{HEVC-EQ}, 80 KBytes by \textit{Liu et al.}, and 86 KBytes by \textit{SC-SKV}, respectively.}
	\label{fig_compareFlowers}
\end{figure}

\begin{figure}[!t]
	\centering
	\includegraphics[width=5.4in]{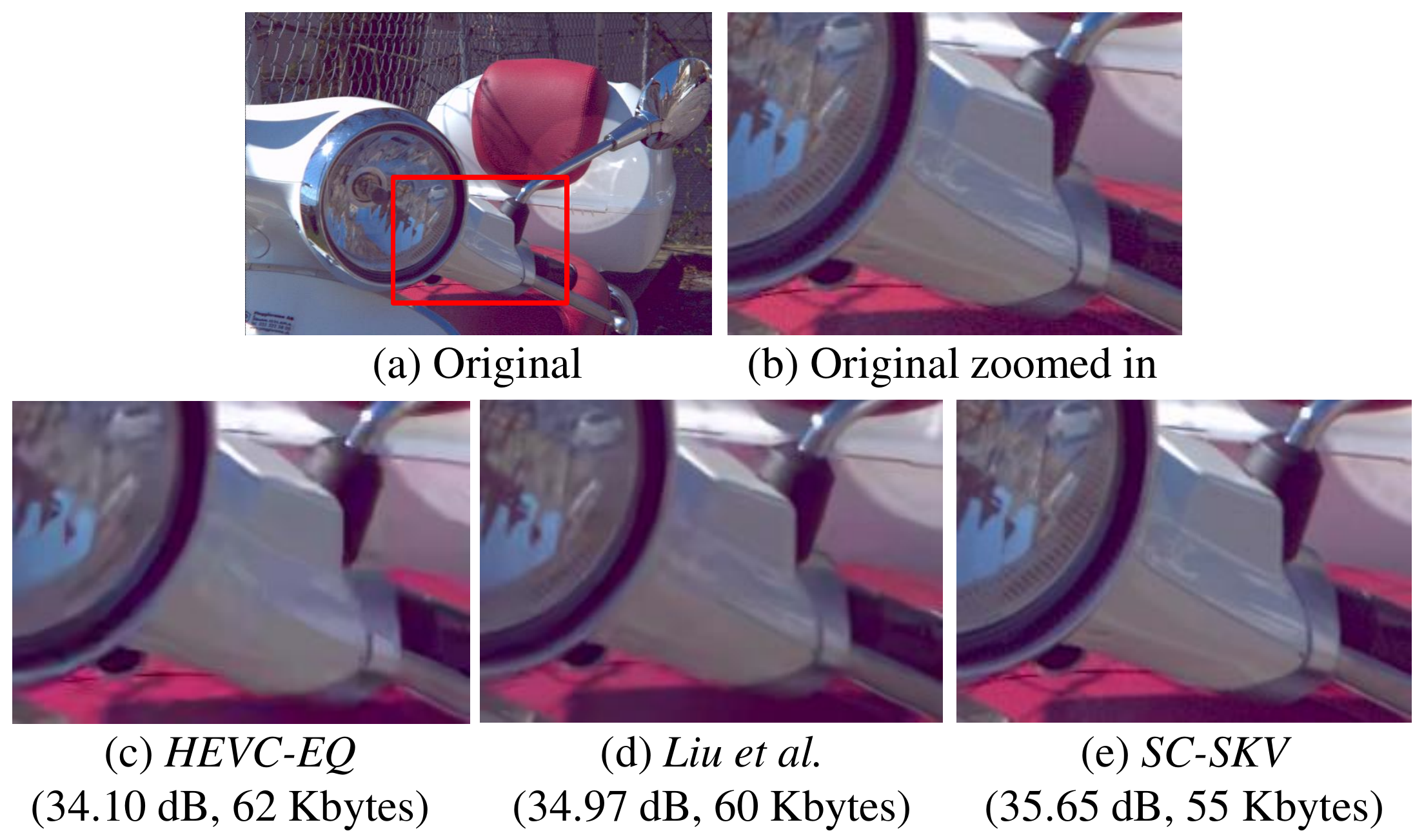}
	\caption{Visual comparison of decoded central view for the LF \textit{I05 Vespa} from different codecs. (b) is the original zoom-in on the red rectangle in (a). (c), (d), (e) are the zoom-in LF central views decoded from compressed bit stream of 62 KBytes by \textit{HEVC-EQ}, 60 KBytes by \textit{Liu et al.}, and 55 KBytes by \textit{SC-SKV}, respectively.}
	\label{fig_compareVespa}
\end{figure}

\begin{figure}[!t]
	\centering
	\includegraphics[width=5.4in]{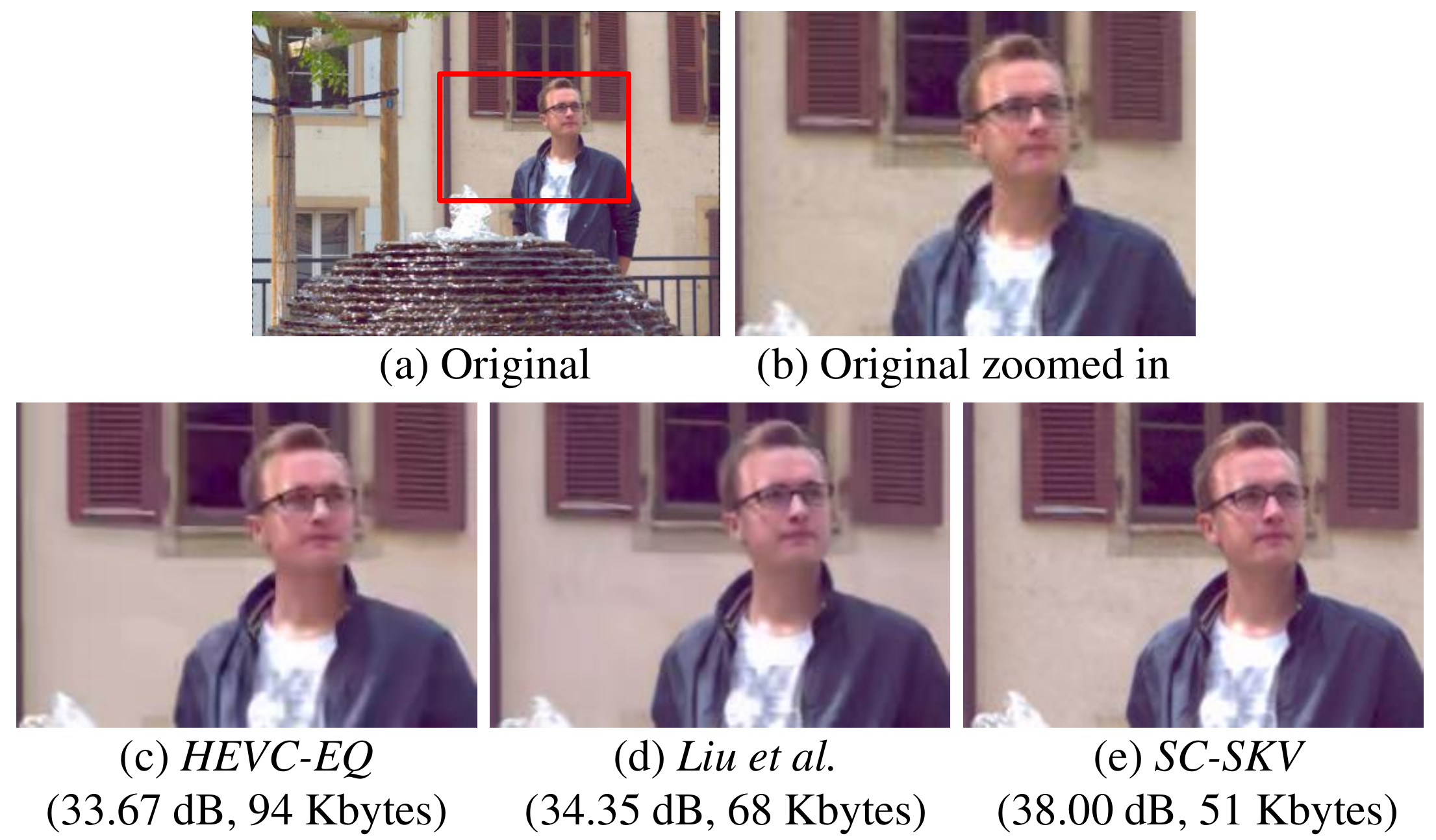}
	\caption{Visual comparison of decoded central view for the LF \textit{I09 Fountain} from different codecs. (b) is the original zoom-in on the red rectangle in (a). (c), (d), (e) are the zoom-in LF central views decoded from compressed bit stream of 94 KBytes by \textit{HEVC-EQ}, 68 KBytes by \textit{Liu et al.}, and 51 KBytes by \textit{SC-SKV}, respectively.}
	\label{fig_compareFountain}
\end{figure}

Visual comparison is carried out at similar compression bit rates for the LF data \textit{I01 Bikes}, \textit{I03 Flowers}, \textit{I05 Vespa}, and \textit{I09 Fountain}, as shown in Fig. \ref{fig_compareBikes}, Fig. \ref{fig_compareFlowers}, Fig. \ref{fig_compareVespa}, and Fig. \ref{fig_compareFountain}, respectively. As can be seen, the visual quality is significantly better for \textit{SC-SKV} codec at low bit rate cases, where reconstructions from \textit{HEVC-EQ} and \textit{Liu et al.} appear blurry and noisy; however the \textit{SC-SKV} codec's results are much smoother in texture-less regions, and much sharper on the edges.

\section{Conclusion} \label{sec_conclusion}

In this paper, we have proposed a LF codec that fully exploits the intrinsic geometry between the LF sub-views by first approximating the LF with disparity guided sparse coding over a perspective shifted light field dictionary. The sparse coding is only based on several optimized Structural Key Views (SKV); yet the entire LF can be recovered from the coding coefficients. By keeping the approximation identical between encoder and decoder, only the residuals of non-key views, the disparity map, and the SKVs need to be saved into bit steam. An optimized SKV selection method has been proposed such that most LF spatial information could be preserved. And to achieve optimum dictionary efficiency, the LF is divided into several Coding Regions (CR), over which the reconstruction works individually. Experiments and comparisons have been carried out over benchmark LF dataset which show that the proposed \textit{SC-SKV} codec produces convincing compression results in terms of both rate-distortion performance and visual quality compared with \textit{High Efficiency Video Coding} (\textit{HEVC}): with 47.87\% BD-rate reduction and 1.59 dB BD-PSNR improvement achieved on average, especially with up to 4 dB improvement for low bit rate scenarios.

\section*{Acknowledgment}
\addcontentsline{toc}{section}{Acknowledgment}
The authors would like to thank Dr. Dong Liu and Mr. Shengyang Zhao for providing their LF compression code \cite{liu2016pseudo} for comparison.


\end{document}